\documentclass[review]{elsarticle}
\graphicspath{ {./figures/} }
\usepackage{hyperref}
\usepackage{float}
\usepackage{verbatim} 
\usepackage{apalike}
\usepackage{amsmath}
\usepackage{multirow}
\usepackage[table,xcdraw]{xcolor}
\restylefloat{figure}
\restylefloat{table}

\journal{Expert Systems with Applications}

\biboptions{authoryear}

\begin{document}
\begin{frontmatter}

\title{A High-Performance Customer Churn Prediction System based on Self-Attention}

\author[label1]{Haotian Wu\corref{cor1}}
\ead{19711032@bjtu.edu.cn}

\cortext[cor1]{Corresponding author.}
\address[label1]{Beijing Jiaotong University, School Economics and Management, Beijing, China}

\begin{abstract}
Customer churn prediction is a challenging domain of research that contributes to customer retention strategy. The predictive performance of existing machine learning models, which are often adopted by churn communities, appear to be at a bottleneck, partly due to models' poor feature extraction capability. Therefore, a novel algorithm, a hybrid neural network with self-attention enhancement (HNNSAE), is proposed in this paper to improve the efficiency of feature screening and feature extraction, consequently improving the model's predictive performance. This model consists of three main blocks. The first block is the entity embedding layer, which is employed to process the categorical variables transformed into 0-1 code. The second block is the feature extractor, which extracts the significant features through the multi-head self-attention mechanism. In addition, to improve the feature extraction effect, we stack the residual connection neural network on multi-head self-attention modules. The third block is a classifier, which is a three-layer multilayer perceptron. This work conducts experiments on publicly available dataset related to commercial bank customers. The result demonstrates that HNNSAE significantly outperforms the other Individual Machine Learning (IML), Ensemble Machine Learning (EML), and Deep Learning (DL) methods tested in this paper. Furthermore, we compare the performance of the feature extractor proposed in this paper with that of other three feature extractors and find that the method proposed in this paper significantly outperforms other methods. In addition, four hypotheses about model prediction performance and overfitting risk are tested on the publicly available dataset.
\end{abstract}


\end{frontmatter}

\section{Introduction}
\label{introduction}

With the rapid development of the financial industry, the customer resources and scale of commercial banks are expanding rapidly\citep{zhao2019reform}. At the same time, under the downward trend of profit rate, financial disintermediation phenomenon is increasingly obvious\citep{jiang2020build}. Customer demand for customized and high-return products and services is growing. Banks not only need to find new sources of revenue, but also need to shift to a user-centered business model\citep{toloba2020outlook}. As a result, competition in the banking industry has become increasingly fierce, which is not only from traditional banking institutions competing for customer resources, but also from new entrants in banking market. These new entrants are typical of some technology-based enterprises, which rely on technological advantages to carry out financial business and provide more customized services by more efficient insight into customer needs based on data-driven methods. The competition for customer resources by new entrants in these markets should not be ignored\citep{shirazi2019big}. Under the competitive pressure of the financial industry, the risk of customer churn in commercial banks increases. The direct impact of customer loss on banks includes the decline of profits and social reputation. The most significant indirect influence is the weakening of imitation effect further induced by the decline of social reputation, which is very unfavorable for banks to mine potential customers and maintain long-term profitability\citep{hogan2003true}. Therefore, customer retention is an important strategy under the scope of customer relationship management (CRM). In terms of profit, research has shown that every $5\%$ more customers retained by a company can lead to an almost $100\%$ boost in profits\citep{reichheld1990zero}. In terms of cost, the consensus is that it costs a lot more to retain a customer than to find a new one\citep{jianxun2012study}. 

The establishment of customer churn model of commercial banks aims to identify the signals of customer churn in advance, and transmit these signals to the managers, to help the managers timely adopt retention strategies for customer groups with churn risks to reduce the losses of the banks\citep{vafeiadis2015comparison}. Recently, the customer churn prediction problem is a typical dichotomous problem, the classification target is “churn” and “not churn” two categories. Recently, data mining technology has been widely used in the construction of customer churn prediction models, and machine learning-related methods have received a lot of attention from researchers because of its high efficiency\citep{tsai2009customer,umayaparvathi2016survey}. In general, machine learning models for customer churn prediction can be mainly divided into three categories: Individual Machine Learning (IML) model and its improved model, Ensemble Machine Learning (EML) model and its improved model, and Deep Learning (DL) model.

The commonly applied IML model and its improved model include Logistic Regression (LR), Decision Tree (DT)\citep{ballings2012customer}, Support Vector Machine (SVM)\citep{kim2005application}, Bayesian Belief Network\citep{kirui2013predicting}, Hidden Markov model(HMM)\citep{huisheng2020customer}, multi-layer perceptron(MLP)\citep{ismail2015multi,wael2022customer} and so on. The structure of these models is often relatively simple, and under the specific data form of customer churn prediction task, they have high forecasting performance\citep{nath2003customer,hur2005customer,hadden2006churn}. At the same time, most of them have good comprehensibility\citep{verbeke2012new}. These single machine learning models are often used by researchers as a baseline.

To further improve the accuracy and generalization of customer churn prediction model, EML models are introduced to solve this problem. EML is proved to combine the positive aspect of individual machine learning models with different structures to decrease their independent errors\citep{nanni2009experimental}, so it has higher accuracy of predictability compared with the stand-alone model. Boosting\citep{hu2005data,lu2012customer}, Bagging\citep{lariviere2005predicting,anil2008predicting,de2011empirical} and multi-stage\citep{tsai2009customer,de2018new} ensemble learning are the main three types of EML methods that are applied in the churn prediction community. In recent years, with the rapid development of deep learning, models represented by deep neural networks have gradually attracted the attention of the customer churn prediction community. Although the interpretability of deep neural network is poor, deep neural network greatly reduces the workload of artificial feature engineering by virtue of its superior feature of adaptive feature extraction\citep{bilal2016predicting,dalli2022impact}. At the same time, some of the methods used in AI niches are beginning to be migrated to churn warning communities. For example, the vector embedding method in the natural language processing community\citep{cenggoro2021deep}, the Deep Q Network (DQN) model in the reinforcement learning (RL) community\citep{panjasuchat2020applying}, etc. Innovatively, this paper is the first effort to introduce self-attention into the model in the customer churn prediction community. In this paper, a hybrid neural network with self-attention enhancement (HNNSAE) is proposed to extract high-correlated features more efficiently and reduce the impact of low-correlated features on model performance by using self-attention and multi-head attention mechanism. This hybrid model firstly uses entity embedding technology\citep{guo2016entity} to map input features to obtain their representation vectors. Then multi-head self-attention block is utilized to screen the features of these representation vectors, and the weighted variables are input into the multi-layer perceptron for training. Experimental results show that the proposed feature extraction method has better performance than artificial feature engineering and deep neural network adaptive feature extraction mechanism. HNNSAE outperforms the baseline model. As the first work to introduce self-attention into the customer churn problem, this research demonstrates this hybrid model based on self-attention outperforms the baseline model through a number of ablation experiments. This work verifies the following four hypotheses on the customer churn data set in the commercial bank: (1) The sample imbalance will significantly affect the performance of the HNNSAE model. (2) Multi-head attention will increase the risk of overfitting when the number of attention heads increases to a certain level. (3) The interaction between the number of attention heads and whether Synthetic Minority Oversampling Technique (SMOTE) is used to handle sample imbalance will significantly affect the predictive performance of the model and the risk of overfitting. (4) The introduction of entity embedding is conducive to improving the performance of the HNNSAE model, and this improvement will become significant as the number of epochs increases.

\section{Related Works}
\label{relate_work}

\subsection{Individual Machine Learning Methods}
\cite{au2003novel} proposed an evolutionary data mining algorithm (DMEL), which can search possible rule space based on evolution approach and predict customer churn more accurately on test data with different churn rates. \cite{kim2005application} proposed to use SVM to predict customer churn and verified that the performance of SVM was better than that of back propagation neural network (BPN) on a credit card customer churn analysis data set. \cite{zhao2005customer} proposed an improved one-class support vector machine. This model introduces slack variables and regards all data points close enough to the origin as abnormal data points. \cite{kirui2013predicting} introduced Bayesian Belief Network into customer churn prediction for the first time in the telecommunication industry. \cite{ballings2012customer} analyzed the entire customer database of a newspaper company. The performance of LR and DT in user churn prediction was compared, and the effect of time window selection on model performance was researched. \cite{ismail2015multi} built a customer churn prediction model based on MLP, and the accuracy of the model reached $91.28\%$. \cite{huisheng2020customer} proposed to construct the HMM to predict customer churn in the telecommunication industry and verified that this model has a predictive performance superior to LIBLINEAR on a real data set.

\subsection{Ensemble Machine Learning Methods}
Ensemble learning is often considered as the effective solution for many machine learning problems. It improves the performance of the model by combining the prediction output of multiple single weak learners\citep{sagi2018ensemble}. Ensemble learning has been widely applied in customer churn prediction. There are mainly three types of ensemble learning structure used in customer churn community: Bagging, Boosting, and Multi-stage method.

In terms of bagging, this paper divides bagging methods into homogeneous bagging and heterogeneous bagging according to whether single weak classifiers are similar or not. Good examples of homogeneous bagging are random forest and its improved algorithms.\cite{anil2008predicting} proposed to build a customer churn prediction model for financial service companies based on random forest. Compared with logistic regression model, random forest has achieved better performance. Aiming to handle the problem of class imbalance, \cite{burez2009handling} proposed to construct weighted random forest to predict customer churn. They performed experiments to verify that this improved random forest algorithm outperforms ordinary random forest when there is an imbalance in the sample. In addition, \cite{de2011empirical} compared the performance of two ensemble classifiers based on rotation, rotation forest and RotBoost, on four customer turnover datasets. It was found that the rotation forest based on independent component analysis (ICA) had good performance and the best AUC was about 0.838. Heterogeneous Bagging is mainly based on a variety of single model voting methods. \cite{anil2008predicting} developed an ensemble system incorporating majority voting which involves MLP, LR, DT, Random Forest, Radial Basis Function network and SVM. Experiments showed that the integrated voting system has high sensitivity and overall accuracy.

In terms of boosting ensemble machine learning, \cite{jinbo2007application} compared the performance of three different Boosting Schemes (including Real Adaboost, Gentle Adaboost and Modest Adaboost) in customer churn prediction. Boosting method was proved to have a significant improvement in accuracy compared with other single machine learning models. \cite{domingos2021experimental} proposed a customer churn prediction model based on XGBOOST, the AUC of which reached 0.85(which was a relatively high level among the test results of similar data sets).

In terms of multi-stage ensemble learning models. \cite{tsai2009customer} proposed a hybrid neural network composed of ANN and self-organizing maps (SMO). In this model, data are clustered by SOM, and then the results of clustering are utilized for ANN training. The optimal accuracy of this hybrid neural network model on the test set is $93.06\%$. \cite{de2018new} proposed a Logit Leaf Model (LLM) composed of segmentation phase and prediction phase. The segmentation phase is implemented by decision tree and the prediction phase by logistic regression. The performance of this model is better than that of a single decision tree and logistic regression model, while taking into account the comprehensibility.

\subsection{Deep Learning Methods and Reinforcement Learning}
Although ensemble machine learning can effectively improve the prediction performance, parallel ensemble methods like bagging are often criticized for their poor interpretability\citep{meinshausen2010node}. In addition, a common drawback of IML and EML is that they tend to require complex feature engineering. The quality of feature extraction largely depends on the method of feature engineering used by the modeler, so it may be subjective and poor in scalability. These extracted features are applied to the training of machine learning models, which will significantly affect the model performance. These shortcomings have been proved to be mitigated using deep learning methods\citep{diro2018leveraging}. An important method in deep learning is neural network. Deep neural network is able to extract higher level features adaptively by continuously stacking hidden layers, which is beneficial to capture the potential relationship between features and reduce the workload of manual feature engineering\citep{bar2014computational}. In recent years, due to its excellent feature extraction ability, deep learning methods have received more and more attention in related research of customer churn prediction\citep{bilal2016predicting}. \cite{domingos2021experimental} set up multiple ablation experiments to explore the influence of different combinations of hyperparameters on the performance of deep neural network based on the public customer churn data set of commercial banks, and the accuracy rate of the model with the best performance in the experiment was $86.9\%$. \cite{cenggoro2021deep} developed an explicable customer churn prediction model based on the vector embedding method in deep learning, and the F1 score of the model was $81.16\%$. \cite{panjasuchat2020applying} introduced reinforcement learning into the customer churn prediction community for the first time. The focus of their work was the robustness of the model. DQN is applied to train on the customer churn data set, and it is verified that DQN, as an active learner, shows strong robustness compared with other machine learning models when the data pattern changes.

\subsection{Attention Mechanism and Self Attention}
Since 2014, the attention mechanism has been utilized as a method of improving learning. Because of its ability to dynamically manage the information flow and filter the low-relevant stimuli by assigning feature weights, it has been regarded as a fundamental concept for advancing deep neural networks by the deep learning community\citep{correia2021attention}. Transformer, as a breakthrough in the attention mechanism research community, completely replaces RNN with a self-attention mechanism for machine translation tasks, which allows it to better capture global dependencies between input and output. At the same time, the introduction of multi-head attention improves the model’s ability to obtain information from different representation subspaces, and by combining this information, the bias caused by single-head attention can be inhibited\citep{vaswani2017attention}.

\section{Methodology}
\label{methodology}

\subsection{Description of Hybrid Neural Network with Self-attention Enhancement (HNNSAE)}
\begin{figure}[htp]
    \centering
    \includegraphics[width =\linewidth]{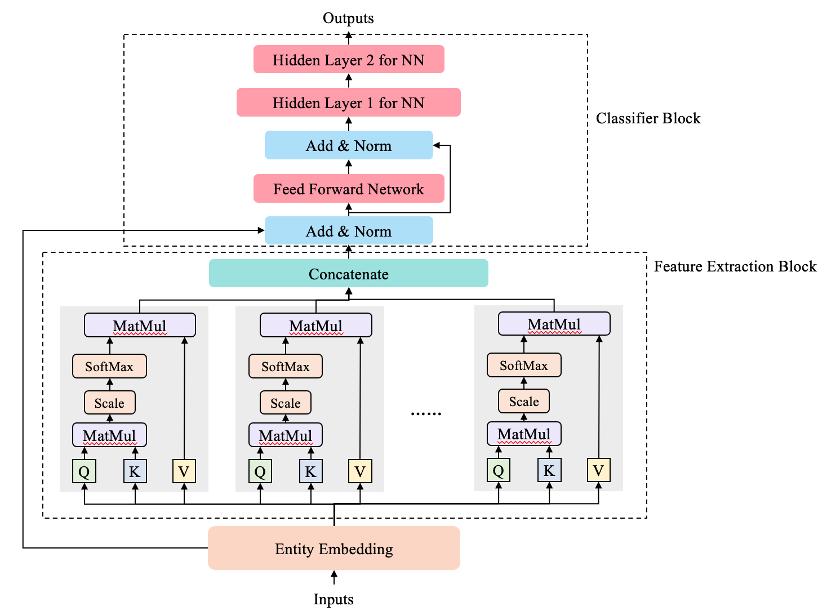}
    \caption{Overall Model Architecture of HNNSAE}
    \label{model}
\end{figure}

Considering the limited number of features in the customer churn data set of commercial banks used in this work, it may be difficult to construct enough effective features to train the model through the traditional artificial feature engineering method. Therefore, the research motivation of this work is to give consideration to the effectiveness and sufficiency of feature extraction. Artificial neural networks can adaptively extract higher-level features by stacking and hiding layers. HNNSAE benefits from its self-attention mechanism, which can effectively capture the global dependency between input and output to achieve feature screening. The overall architecture of HNNSAE model is shown in Figure ~\ref{model}.
Firstly, to concentrate the information of discrete category variables, we employ the entity embedding method to transform them into non-discrete representations. The work of \cite{de2015artificial} have proven the effectiveness of entity embedding in handling discrete structured data. Entity embedding reduces memory usage by mapping discrete structured data to the embedded space and reveals the intrinsic properties of categorical variables through the similarity of the embedded vectors. Secondly, these non-discrete representations are fed to the multi-headed self-attention module. Self-attention block in every head follows the scaled dot-production attention pattern. Scaled dot-production attention is a Query-Key-Value model that calculates the output of Attention matrix using the following equation\citep{lin2021survey}:
\begin{equation}
    \operatorname{Attention}(Q, K, V)=\operatorname{Softmax}\left(\frac{Q K^{T}}{\sqrt{d_{k}}}\right)
\end{equation}
Where $Q$ is the packed matrix representation of queries, $K$ is the packed matrixed representation of keys, $V$ is the packed matrixed representation of values, $d_{k}$ is dimension of keys. In addition, the adoption of multi-head attention mechanism can make the feature filter capture the feature information of different representation subspaces. First, original $Q$, $K$, and $V$ are projected into $H$ different subspaces and scaled dot-production attention was performed, respectively. Then, the outputs are concatenated and re-projected. The specific calculation process is as follows:
\begin{equation}
    \begin{aligned}
        \text { MultiHead }(Q, K, V) & = \text { Concat }\left(\text { head }_{1}, \ldots, \text { head }_{H}\right) W^{O} \\
        \text { where head } & = \text { Attention }\left(Q W_{i}^{Q}, K W_{i}^{K}, V W_{i}^{V}\right)
    \end{aligned}
\end{equation}

In addition, in order to better extract complex features, we stack the residual connection neural network on the multi-head self-attention module. Finally, these processed features are fed into a multi-layer perceptron and classified

\section{Experimental Setup}
\label{experiment}
\subsection{Dataset}
The datasets used in this paper are exposed by the Kaggle platform, which is download from Kaggle\footnote{https://www.kaggle.com/barelydedicated/bank-customer-churn-modeling} on 30 April,2021. It contains 10, 000 pieces of real-world commercial bank customer data with 14 characteristics. The basic description of each column is shown in Table ~\ref{table1}.
\begin{table}[htp]

\caption{Basic description of each column}
\label{table1}
\begin{tabular}{ccc}
\hline
Column          & Data Typy & Description                                                                       \\ \hline
RowNumer        & Int 64    & Row number of customers                                                           \\
CustomerId      & Int 64    & Unique identification of each customer                                            \\
Surname         & String    & Surname of customer                                                               \\
CreditScore     & Int 64    & A score that a bank assigns to a customer based on their historical credit        \\
Geography       & String    & The country where the customer is currently located                               \\
Gender          & String    & The gender of customer                                                            \\
Age             & Int 64    & The age of customer                                                               \\
Tenure          & Int 64    & The number of years the customer account has been with the bank                   \\
Balance         & Float 64  & The balance of the customer’s bank account                                        \\
NumOfProducts   & Int 64    & The number of bank products the customer has used.                                \\
HasCrCard       & Int 64    & Whether the customer has a credit card                                            \\
IsActiveMember  & Int 64    & Whether the customer is active                                                    \\
EstimatedSalary & Float 64  & The estimate of the customer’s current salary                                     \\
Exited          & Int 64    & \begin{tabular}[c]{@{}c@{}}Prediction Target. \\ 1: Churn, 0: Retain\end{tabular} \\ \hline
\end{tabular}
\end{table}

\subsection{Data Preparation}
First, this work deletes “RowNumber”, “CustomerID” and “SurName”, three columns that are not relevant to modeling. Secondly, the data types were divided. In this work, five columns, “Creditsore”, “Age”, “Tenure”, “Balance”, and “EstimateSalary”, were regarded as non-category characteristics. Two string variables, “Geography” and “Gender” are treated as category characteristics and coded as dummy variables. Three numeric variables with fewer unique values, “IsActiveMember”, “HasCrCard”, and “NumOfProducts” , are also considered category characteristics. The third step is to conduct basic descriptive statistical analysis on the non-categorical features and observe the variability, kurtosis, skewness, and other properties of the data. “CreditScore” has the least data variability, while "Balance" has the most. In terms of skewness, the data distribution of the five non-categorical features is close to the Gaussian distribution, so the data processing and research focus of this work is not on the transformation methods that change data distribution. More Details of descriptive statistics for non-categorical feature will be shown in Table~\ref{table2}.
\begin{table}[]
\centering
\caption{Descriptive statistics for non-category variables}
\begin{tabular}{c|ccccc}
\hline
\makebox[0.1\textwidth][c]{Variable Name} & \makebox[0.1\textwidth][c]{Std}           & \makebox[0.1\textwidth][c]{CV} & Kurtosis                       & Mean     & Skewness \\ \hline
CreditScore   & 96.653 & 0.149 & -0.426 & 650.5288 & -0.071   \\
Age             & 10.488     & 0.269 & 1.395  & 38.921     & 1.011  \\
Tenure          & 2.892      & 0.577 & -1.165 & 5.013      & 0.011  \\
Balance         & 62397.4052 & 0.816 & -1.489 & 76485.889  & -0.141 \\
EstimatedSalary & 57510.493  & 0.575 & -1.182 & 100090.240 & 0.002  \\ \hline
\end{tabular}
\label{table2}
\end{table}

\begin{figure}
    \centering
    \includegraphics[width=0.6\linewidth]{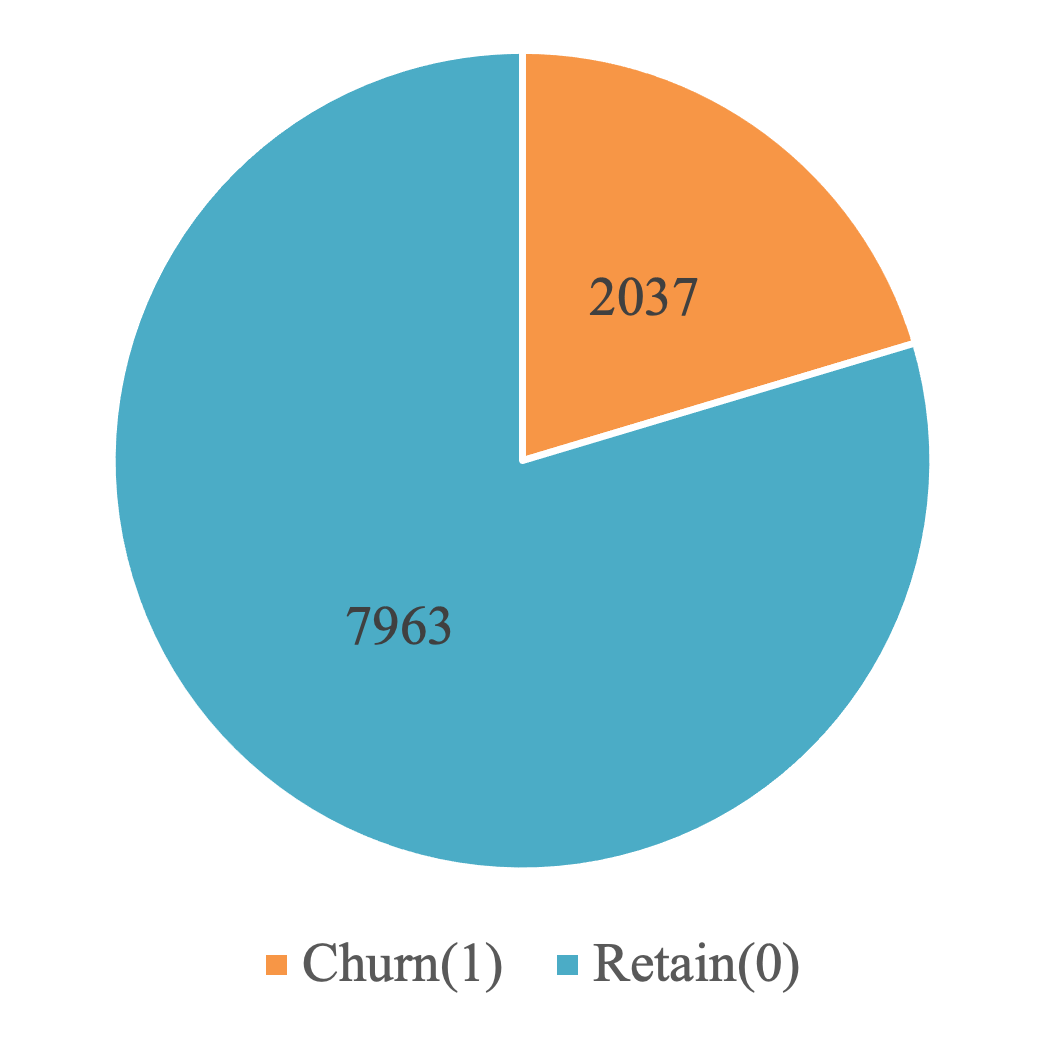}
    \caption{Proportional visualizations for the two categories in “Exited”}
    \label{figure2}
\end{figure}

In addition, considering that the sample imbalance may have a negative impact on the model performance, this work conducted a proportional visualization analysis of the two categories of "Exited", and the visualization results were shown in Figure ~\ref{figure2} Through observation, it can be found that the number of samples labeled “0” is close to four times of the number of samples labeled “1”, and the sample is obviously imbalanced. Therefore, this work is based on SMOTE to process the imbalanced samples. At the same time, the ablation experiment was set to compare the prediction performance of the model before and after the introduction of SMOTE method. SMOTE is an improved method based on random over-sampling algorithm, which can synthesize new samples according to minority class and add them to the data set \citep{chawla2002smote}

In this work, the data set was split into the train set ($80\%$ of data) and test set ($20\%$ of data). Meanwhile, aiming to eliminate the contingency of experimental results, each ablation experiment was carried out five times and the data set of each experiment was shuffled to obtain different train sets and test sets.

\subsection{Baselines and Training Specifications}
The main purposes of this research are as follows:

\textbf{(1).Verify whether the sample imbalance will significantly affect the performance of the HNNSAE model.} 

This work sets up two sets of ablation experiments, each of which is carried out for five times, and the data used in each experiment is randomly shuffled. In both experiments, 8-heads attention and three-layer neural network are used, and the hidden layers of the neural network are (128,64,32). Adam optimizer is adopted. Each model is trained 1000 epochs, and experimental data (train loss and test AUC) are collected at the 500th epoch and the 1000th epoch. The difference is that the first group of experiments don’t handle the sample imbalance, while SMOTE is used to handle the sample imbalance in the second group. The experimental results are shown in session ~\ref{5.1}.

\textbf{(2). Verify whether multi-head attention will increase the risk of overfitting when the number of attention heads increases to a certain level.} (Some research have suggested that only a small part of attention heads contribute to the final prediction, and they tested this hypothesis in ablation experiments using the Attention Head Pruning approach\citep{hao2021self}.)

In this work, four groups of ablation experiments are set up, and each group of ablation experiments is carried out five times. The number of attention heads used in the four groups of ablation experiments are 2, 4, 8, 16, and the remaining the network hyperparameters are kept consistent. The four groups of experiments all use SMOTE to deal with sample imbalance. Train loss and test AUC are collected every 50 epochs. Calculate ROD and ROIn at the 500th epoch and the 1000th epoch, respectively. The experimental results are shown in session ~\ref{5.2}.

\textbf{(3). Verify whether the interaction between the number of attention heads and whether SMOTE is used to handle sample imbalance will significantly affect the predictive performance of the model and the risk of overfitting.} 

In this work, based on the four ablation experiments in (2), a similar four ablation experiments are performed without processing the data with imbalanced samples. Train loss and test AUC are collected and ROD and ROIn are calculated at 500th epoch and 1000th epoch. Then two-factors analysis of variance (ANOVA) is used to verify the influence of the interaction between the number of attention heads and sample imbalance on the prediction performance of the model. The experimental results are shown in session ~\ref{5.3}.

\textbf{(4). Verify whether the introduction of Entity Embedding is conducive to improving the performance of the HNNSAE model, and this improvement will become significant as the epoch increases.} 

In this work, two groups of ablation experiments are set up, and each group of ablation experiments is carried out five times. One group retains entity embedding while the other group removes it, and the remaining networks’ hyperparameter are kept consistent. SMOTE is used in both groups to deal with the sample imbalance. Train loss and test AUC are collected every 50 epochs. At the same time, right-tailed tests are performed on train loss and test AUC under the same epoch of two groups of ablation experiments for 20 times in total. The experimental results are shown in session ~\ref{5.4}.

\textbf{(5). Verify whether HNNSAE outperforms the basic models.
In this work, seven groups of ablation experiments are set up, and each group of ablation experiments is carried out five times.}

This paper adopts several models commonly available in the existing literature as baselines, including LR, DT, random forests (RF), XGBOOST, GBDT, ANN, DeepFM+ANN, GBDT+ANN, Autoecnoder+ANN. This paper conducts the one-tailed test for these 9 baselines test AUC and HNNSAE respectively. The experimental results are shown in session ~\ref{5.5}.

\subsection{Evaluation Metrics}
Area Under Curve (AUC) is the main metrics used to evaluate model predictive performance in this work. It represents the area under the Receiver Operating Characteristic Curve (ROC). AUC doesn’t thresholds and can report each type of error collective synthetically\citep{xie2009customer}. Therefore, AUC is generally considered to be a better overall evaluation metrics than accuracy\citep{langley2000crafting}. AUC can be calculated by the following equation:
\begin{equation}
    A U C=\frac{S-\frac{M \times(M+1)}{2}}{M \times N}
\end{equation}
Where $S$ is the sum of score rank of Positive sample, $M$ and $N$ represent the number of positive and negative samples, respectively\citep{huang2012customer}.

Aiming to evaluate the performance improvement rate of the model on the train set and test set and analyze the over-fitting problem, this paper proposes two series of metrics, Rate of decrease of train loss($ROD$) and Rate of increase of test AUC($ROI_{n}$). These two metrics are relative values, which eliminate the dimensional effect to a certain extent and are easy to compare. The calculation formulas of metrics under these two series appeared in the paper are as follows:(The subscript represents the epoch number)
\begin{equation}
    R O D_{1}=\frac{Train\;Loss_{50} - Train\;Loss_{500}}{Train\; Loss_{50}}
\end{equation}
\begin{equation}
    R O D_{2}=\frac{Train\;Loss_{500} - Train\;Loss_{1000}}{Train\; Loss_{500}}
\end{equation}
\begin{equation}
    R O In_{1}=\frac{Test\;AUC_{500} - Test\;AUC_{50}}{Test\; AUC_{50}}
\end{equation}
\begin{equation}
    R O In_{2}=\frac{Test\;AUC_{1000} - Test\;AUC_{500}}{Test\; AUC_{500}}
\end{equation}
\section{Experimental Results and Analysis}

\subsection{Synthetic Minority Oversampling Technique}
\label{5.1}
From the experimental results, the sample imbalance has a great impact on the model prediction performance, which reflects the two aspects of the test AUC and the stability of the test AUC. According to the results of five experiments, the training effect and predictive performance of HNNSAE have been significantly improved after SMOTE data processing. According to the average value of Test AUC, at the 500th epoch and 1000th epoch, after SMOTE treatment, HNNSAE’s test AUC averages 0.9394 and 0.9438, respectively. Std of test AUC of HNNSAE trained on the data that is treated with SMOTE are significantly lower than that without SMOTE treatment, indicating that the predictive performance of HNNSAE on the data that is treated with SMOTE is more stable. The specific experimental results are shown in Table ~\ref{table3}:
\begin{table}[htp]
\centering
\caption{Predictive performance before and after SMOTE}
\label{table3}
\begin{tabular}{c|ccc}
\hline
Epochs                & Processing   & \begin{tabular}[c]{@{}c@{}}Train Loss\\ ($mean \pm std$)\end{tabular} & \begin{tabular}[c]{@{}c@{}}Test AUC\\ ($mean \pm std$)\end{tabular} \\ \hline
\multirow{2}{*}{500}  & Before SMOTE & $569.179 \pm 16.577$                                            & $0.808 \pm 0.007$                                            \\
                      & After SMOTE  & $50.754 \pm 4.569$                                              & $0.940 \pm 0.0010$                                           \\ \hline
\multirow{2}{*}{1000} & Before SOMTE & $534.887 \pm 12.165$                                            & $0.788 \pm 0.006$                                           \\
                      & After SOMTE  & $38.704 \pm 5.971$                                             & $0.944 \pm 0.002$                                           \\ \hline
\end{tabular}
\end{table}
According to the experimental results, we also observed that the test AUC at the time of $1000 \;th$ epoch without SMOTE treatment was significantly lower than that at the time of $500 \;th$ epoch, while the train loss decreased during the process of $500\; th$ epoch $\sim$ $1000\; th$ epoch. This indicates that the HNNSAE model overfits under this circumstance. When the samples are imbalanced, the model does not learn enough from the minority class but depends more on the majority class, which may increase the risk of overfitting of the model.

\subsection{Multi-head Self-Attention}
\label{5.2}

\begin{figure}[htp]
    \centering
    \includegraphics[width=\linewidth]{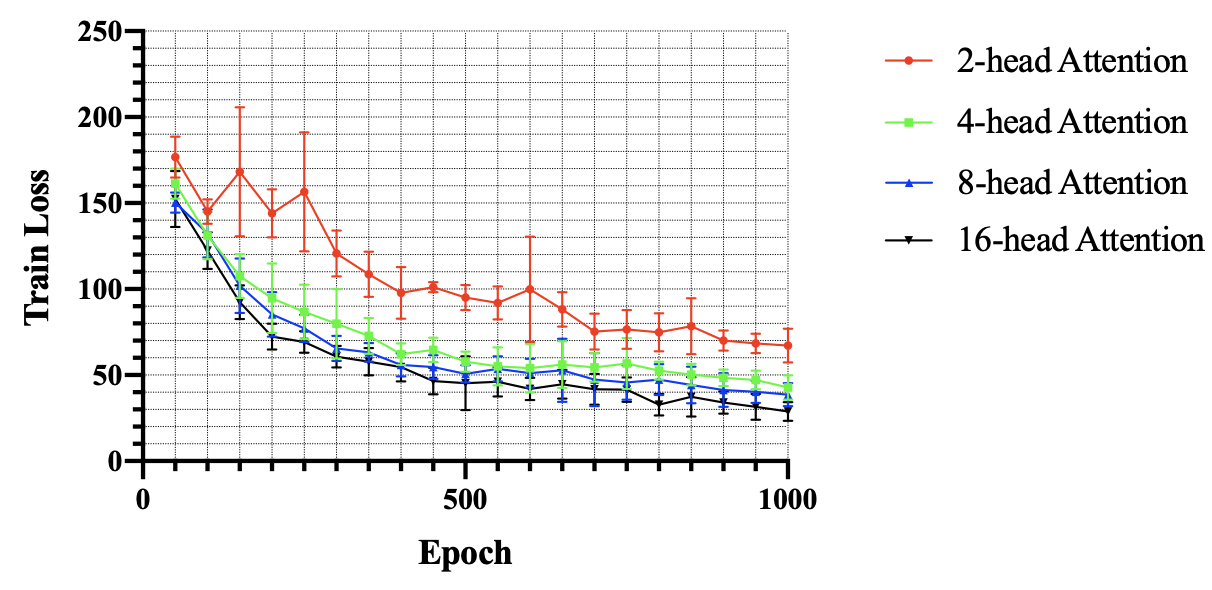}
    \caption{Train loss for HNNSAE with different number of attention heads.}
    \label{figure3}
\end{figure}

\begin{figure}[htp]
    \centering
    \includegraphics[width=\linewidth]{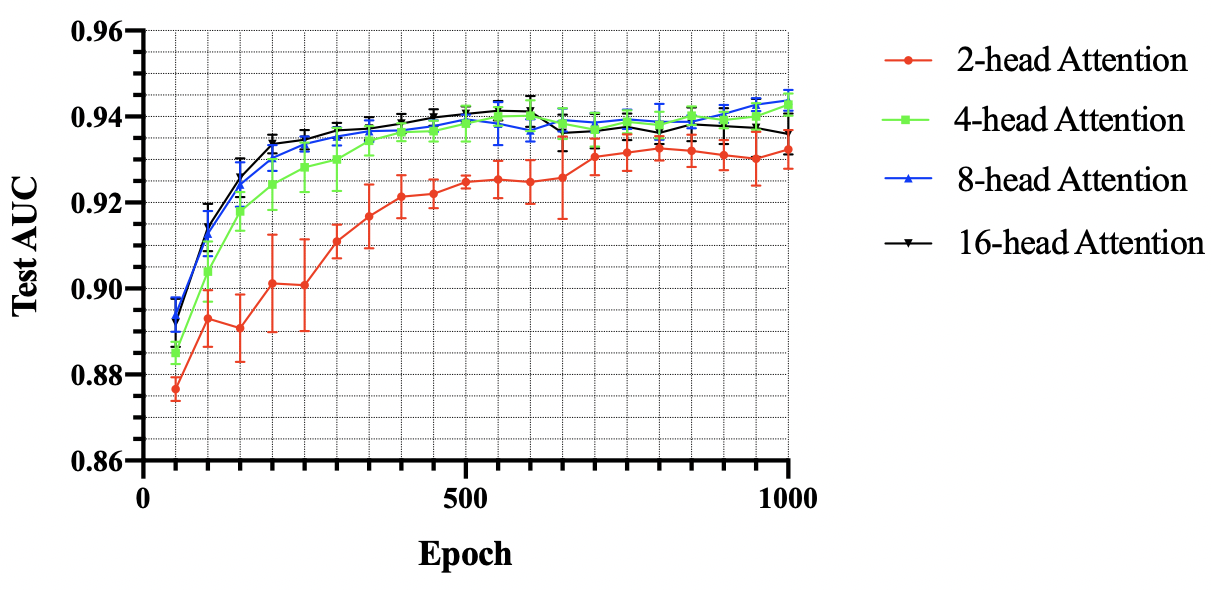}
    \caption{Test AUC for HNNSAE with different number of attention heads.}
    \label{figure4}
\end{figure}

By observing the changes of train loss and test $AUC$ in ablation experiments with four groups of different attention heads, we find that with the increase in the number of attention heads, train loss decreases faster and reaches a lower level. As the number of attention heads increases, the test $AUC$ reaches a relatively stable level more quickly. However, we also find that with the increase of epochs, the test $AUC$ of 16-head self-attention HNNSAE is slightly lower than that of 8-head self-attention and 4-head attention HNNSAE. Although 16-head self-attention HNNSAE is lower in train loss than 8-head self-attention and 4-head self-attention, the test $AUC$ always improve predictive performance of  HNNSAE, and to some extent, it increases the risk of model overfitting. In this study, $ROD$ and $ROIn$ are introduced to evaluate the increasing rate of test $AUC$ and decreasing rate of train loss, so as to further verify whether the increase of the number of attention heads will increase over-fitting risk of model.

\begin{figure}[htp]
\centering
\includegraphics[width=0.8\linewidth]{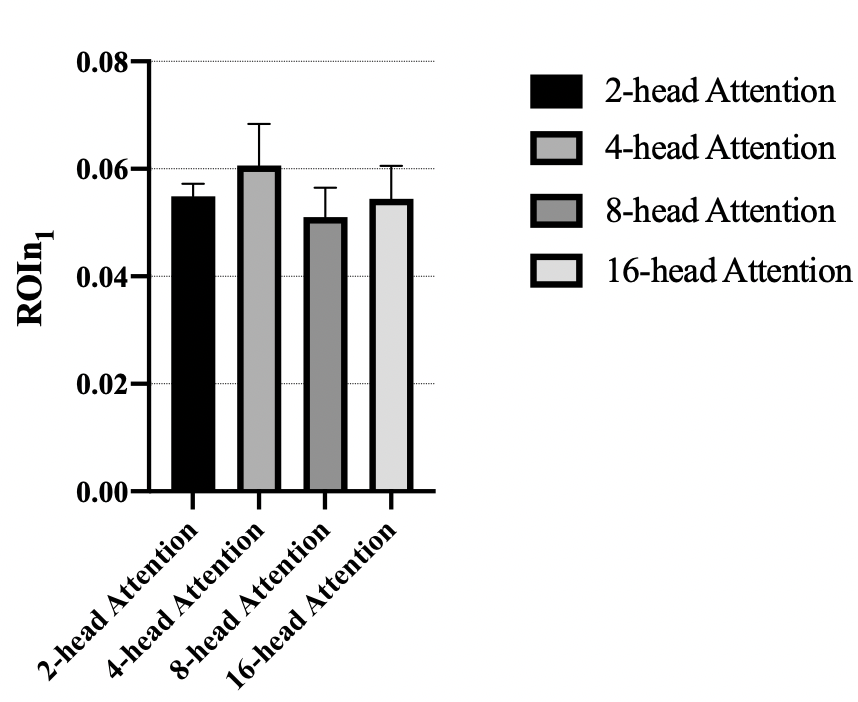}
\caption{ROIn during $50th \sim 500th$ epoch}
\end{figure}

\begin{figure}[htp]
\centering
\includegraphics[width=0.8\linewidth]{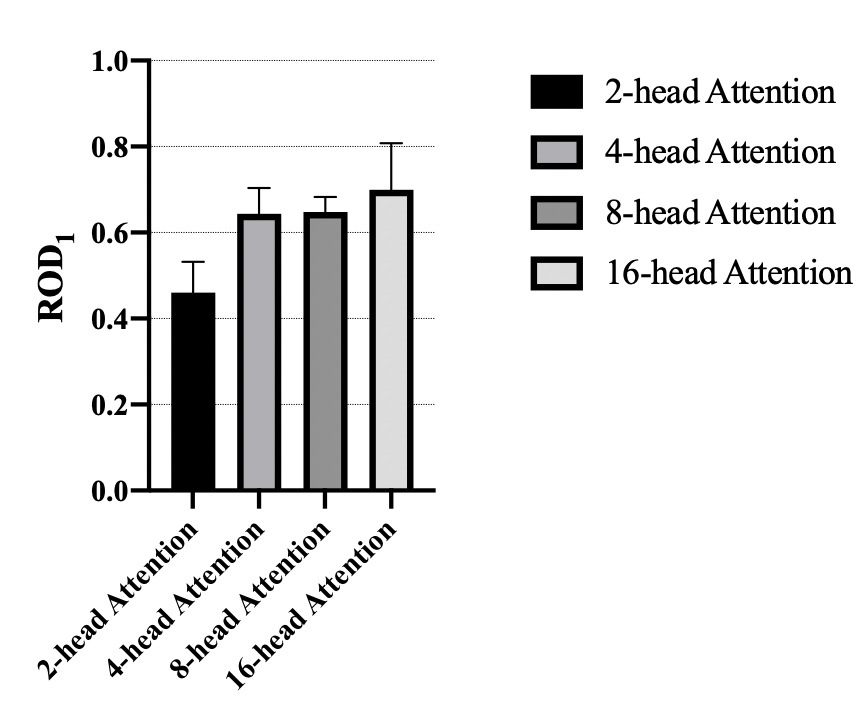}
\caption{ROD during $50th \sim 500th$ epoch.}
\end{figure}

\begin{figure}[htp]
\centering
\includegraphics[width=0.8\linewidth]{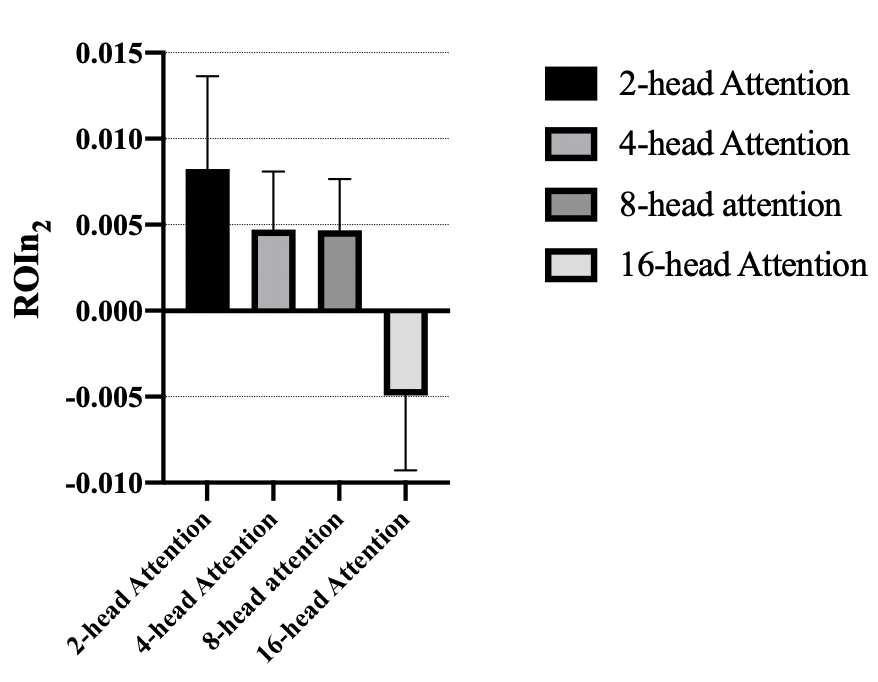}
\caption{ROIn during $500th \sim 1000th$ epoch.}
\end{figure}

\begin{figure}[htp]
\centering
\includegraphics[width=0.8\linewidth]{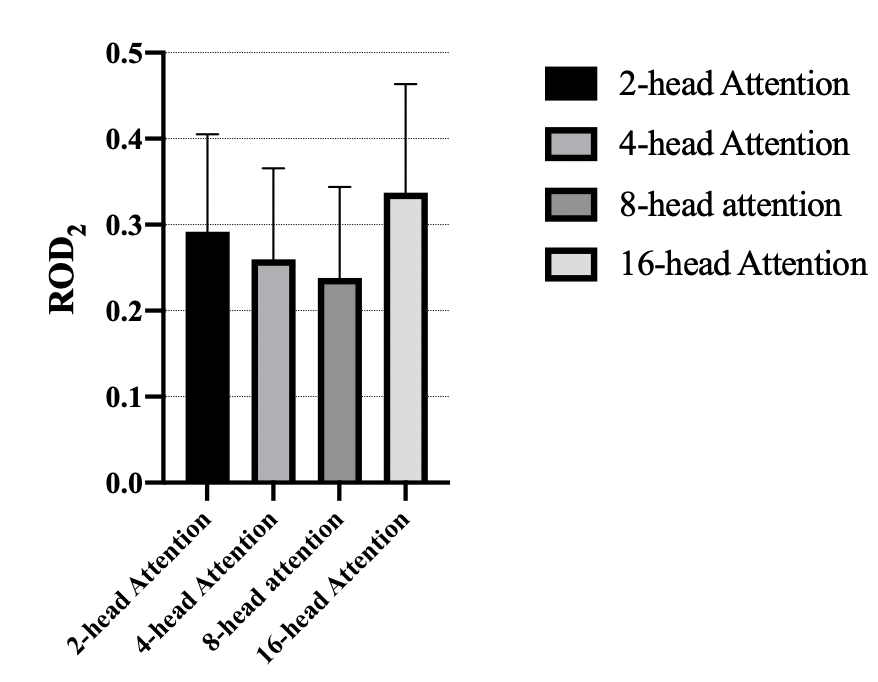}
\caption{ROD during $500th \sim 1000th$ epoch.}
\end{figure}

We observe that the ROD of 16-head self-attention HNNSAE is the largest during $500\;th \sim 1000\;th$ epoch, but the $ROIn$ decreases to a negative value. The $ROIn$ of the other three groups of experimental models during $500\;th \sim 1000\;th$ epoch decreased compared with that during $50\;th \sim 500\;th$ epoch, but still remaines positive. 2-head self-attention HNNSAE has the largest ROIn during $500\;th \sim 1000\;th$ epoch. Although the test AUC of 2-head self-attention HNNSAE is lower than that of the other three models, it has a smaller risk of overfitting.

\subsection{Multi-head Self-Attention, Sample Imbalance and Overfit}
\label{5.3}
In this part, we will further verify whether the increase of the number of attention heads increases risk of over-fitting. In addition, we introduce the ablation experiment of whether SMOTE is used to handle the sample imbalance and verify whether the interaction between the number of attention heads and whether SMOTE is used to handle sample imbalance will significantly affect the predictive performance of the model and the risk of overfitting through two-factor analysis of variance. Table ~\ref{table4} and Table ~\ref{table5} show the predictive performance of HNNSAE with different number of attention heads under balanced and imbalanced samples when the 500th epoch and the 1000th epoch. Table ~\ref{table6} to Table ~\ref{table9} show the results of two-factor analysis of variance.
\begin{table}[htp]
\centering
\caption{Predictive performance with handling sample imbalance}
\label{table4}
\begin{tabular}{c|ccccc}
\hline
Epochs &
  Attention &
  \begin{tabular}[c]{@{}c@{}}Train Loss\\ ($mean\pm std$)\end{tabular} &
  \begin{tabular}[c]{@{}c@{}}ROD\\ ($mean\pm std$)\end{tabular} &
  \begin{tabular}[c]{@{}c@{}}Test AUC\\ ($mean\pm std$)\end{tabular} &
  \begin{tabular}[c]{@{}c@{}}ROI\\ ($mean\pm std$)\end{tabular} \\ \hline
\multirow{4}{*}{500}  & \begin{tabular}[c]{@{}c@{}}2-heads\\ Attention\end{tabular}  & $95.003\pm 6.576$          & $0.459\pm 0.067$ & $0.925\pm 0.0013$ & $0.055\pm 0.0390$  \\
                      & \begin{tabular}[c]{@{}c@{}}4-heads \\ Attention\end{tabular} & $57.776\pm 5.320$   & $0.644\pm 0.054$ & $0.938\pm 0.0038$ & $0.061\pm 0.0070$  \\
                      & \begin{tabular}[c]{@{}c@{}}8-heads \\ Attention\end{tabular} & $50.754\pm 4.569$   & $0.648\pm 0.031$ & $0.939\pm 0.0010$ & $0.051\pm 0.0049$  \\
                      & \begin{tabular}[c]{@{}c@{}}16-heads\\ Attention\end{tabular} & $45.337\pm 13.9061$ & $0.700\pm 0.097$ & $0.941\pm 0.0016$ & $0.054\pm 0.0055$  \\ \hline
\multirow{4}{*}{1000} & \begin{tabular}[c]{@{}c@{}}2-heads\\ Attention\end{tabular}  & $69.184\pm 8.443$   & $0.270\pm 0.097$ & $0.932\pm 0.0040$ & $0.008\pm 0.0051$  \\
                      & \begin{tabular}[c]{@{}c@{}}4-heads\\ Attention\end{tabular}  & $42.776\pm 6.442$   & $0.260\pm 0.095$ & $0.943\pm 0.0023$ & $0.005\pm 0.0030$  \\
                      & \begin{tabular}[c]{@{}c@{}}8-heads \\ Attention\end{tabular} & $38.704\pm 5.971$   & $0.238\pm 0.095$ & $0.944\pm 0.0021$ & $0.005\pm 0.0027$  \\
                      & \begin{tabular}[c]{@{}c@{}}16-heads\\ Attention\end{tabular} & $28.821\pm 4.958$   & $0.377\pm 0.113$ & $0.936\pm 0.0043$ & $-0.005\pm 0.0039$ \\ \hline
\end{tabular}
\end{table}
\begin{table}[htp]
\centering
\caption{Predictive performance without handling sample imbalance}
\label{table5}
\begin{tabular}{c|ccccc}
\hline
Epochs &
  Attention &
  \begin{tabular}[c]{@{}c@{}}Train Loss\\ ($mean\pm std$)\end{tabular} &
  \begin{tabular}[c]{@{}c@{}}ROD\\ ($mean\pm std$)\end{tabular} &
  \begin{tabular}[c]{@{}c@{}}Test AUC\\ ($mean\pm std$)\end{tabular} &
  \begin{tabular}[c]{@{}c@{}}ROI\\ ($mean\pm std$)\end{tabular} \\ \hline
\multirow{4}{*}{500}  & \begin{tabular}[c]{@{}c@{}}2-heads\\ Attention\end{tabular}  & $541.851\pm 4.860$  & $0.247\pm 0.023$ & $0.803\pm 0.0037$ & $-0.023\pm 0.0135$ \\
                      & \begin{tabular}[c]{@{}c@{}}4-heads \\ Attention\end{tabular} & $589.040\pm 20.671$ & $0.219\pm 0.025$ & $0.813\pm 0.0094$ & $0.005\pm 0.0164$  \\
                      & \begin{tabular}[c]{@{}c@{}}8-heads \\ Attention\end{tabular} & $512.334\pm 15.072$ & $0.286\pm 0.021$ & $0.797\pm 0.0093$ & $-0.031\pm 0.0105$ \\
                      & \begin{tabular}[c]{@{}c@{}}16-heads\\ Attention\end{tabular} & $473.904\pm 11.909$ & $0.335\pm 0.019$ & $0.792\pm 0.0085$ & $-0.042\pm 0.0135$ \\ \hline
\multirow{4}{*}{1000} & \begin{tabular}[c]{@{}c@{}}2-heads\\ Attention\end{tabular}  & $447.540\pm 5.131$  & $0.270\pm 0.097$ & $0.782\pm 0.0043$ & $-0.027\pm 0.0070$ \\
                      & \begin{tabular}[c]{@{}c@{}}4-heads\\ Attention\end{tabular}  & $522.847\pm 23.227$ & $0.113\pm 0.013$ & $0.795\pm 0.0108$ & $-0.022\pm 0.0087$ \\
                      & \begin{tabular}[c]{@{}c@{}}8-heads \\ Attention\end{tabular} & $405.619\pm 10.830$ & $0.208\pm 0.025$ & $0.773\pm 0.0090$ & $-0.029\pm 0.0137$ \\
                      & \begin{tabular}[c]{@{}c@{}}16-heads\\ Attention\end{tabular} & $368.428\pm 16.600$ & $0.223\pm 0.024$ & $0.769\pm 0.0065$ & $-0.029\pm 0.0097$ \\ \hline
\end{tabular}
\end{table}

\begin{table}[htp]
\centering
\caption{ANOVA table for test AUC (500 epochs)}
\label{table6}
\begin{tabular}{lllllll}
\hline
Source of Variance & SS     & df & MS         & F         & P-value             & F crit \\ \hline
Factor A           & 0.1809 & 1  & 0.1809     & 4124.3089 & \textbf{2.1411E-35} & 4.1491 \\
Factor B           & 0.0008 & 3  & 0.0003     & 5.9649    & \textbf{0.0024}     & 2.9011 \\
Interaction        & 0.0013 & 3  & 0.0004     & 10.0838   & \textbf{7.8841E-05} & 2.9011 \\
Error              & 0.0014 & 32 & 4.3863E-05 &           &                     &        \\
Total              & 0.1844 & 39 &            &           &                     &        \\ \hline
\end{tabular}
\end{table}

\begin{table}[htp]
\centering
\caption{ANOVA table for test AUC (1000 epochs)}
\label{table7}
\begin{tabular}{lllllll}
\hline
Source of Variance & SS     & df & MS         & F         & P-value             & F crit \\ \hline
Factor A           & 0.2525 & 1  & 0.2525     & 5373.6015 & \textbf{3.1921E-37} & 4.1491 \\
Factor B           & 0.0015 & 3  & 0.0005     & 10.3332   & \textbf{6.5483E-05} & 2.9011 \\
Interaction        & 0.0010 & 3  & 0.0003     & 6.9955    & \textbf{0.0009}     & 2.9011 \\
Error              & 0.0015 & 32 & 4.6987E-05 &           &                     &        \\
Total              & 0.2564 & 39 &            &           &                     &        \\ \hline
\end{tabular}
\end{table}
\begin{table}[htp]
\caption{ANOVA table for test ROIn (500 epochs)}
\label{table8}
\begin{tabular}{lllllll}
\hline
Source of Variance & SS     & df & MS     & F        & P-value             & F crit \\ \hline
Factor A           & 0.0605 & 1  & 0.0605 & 455.3575 & \textbf{1.7245E-20} & 4.1491 \\
Factor B           & 0.0041 & 3  & 0.0014 & 10.3010  & \textbf{6.7063E-05} & 2.9011 \\
Interaction        & 0.0021 & 3  & 0.0007 & 5.3128   & \textbf{0.0044}     & 2.9011 \\
Error              & 0.0043 & 32 & 0.0001 &          &                     &        \\
Total              & 0.0710 & 39 &        &          &                     &        \\ \hline
\end{tabular}
\end{table}
\begin{table}[htp]
\caption{ANOVA table for test ROIn (1000 epochs)}
\label{table9}
\begin{tabular}{lllllll}
\hline
Source of Variance & SS     & df & MS         & F        & P-value             & F crit \\ \hline
Factor A           & 0.0089 & 1  & 0.0089     & 121.9935 & \textbf{1.8864E-12} & 4.1491 \\
Factor B           & 0.0004 & 3  & 0.0001     & 1.7757   & \textbf{0.1716}     & 2.9011 \\
Interaction        & 0.0002 & 3  & 7.4092E-05 & 1.0212   & \textbf{0.3962}     & 2.9011 \\
Error              & 0.0023 & 32 & 7.255E-05  &          &                     &        \\
Total              & 0.0118 & 39 &            &          &                     &        \\ \hline
\end{tabular}
\end{table}

Note: Factor A: Whether SMOTE is used to handle sample imbalance. Factor B: The number of attention heads

In terms of test AUC, whether SMOTE is used to handle sample imbalance has the most significant impact on model predictive performance, and its significance increases with increase of epoch number. The number of attention heads has a relatively significant impact on predictive performance of model, and its significance increases with increase of the epoch number. The interaction between the two factors also has a significant influence on the predictive performance of the model, but the significance decreases with the increase of epoch number. In terms of ROIn, whether SMOTE is used to handle sample imbalance has the most significant impact on the overfitting risk of the model, and the significance decreases with the increase of epoch number. When The number of epochs is 500, The influence of the number of attention heads on overfitting risk of model prediction is significant, but with the increase of the number of epochs, The significance is greatly weakened. At the 500th epoch, the interaction between the two factors had a significant impact on the overfitting risk of the model at the $95\%$ confidence level, but this significance was significantly weakened as the epochs number increased to 1000.

\subsection{The Necessity of Entity Embedding}
\label{5.4}
\begin{figure}[htp]
    \centering
    \includegraphics[width=0.9\linewidth]{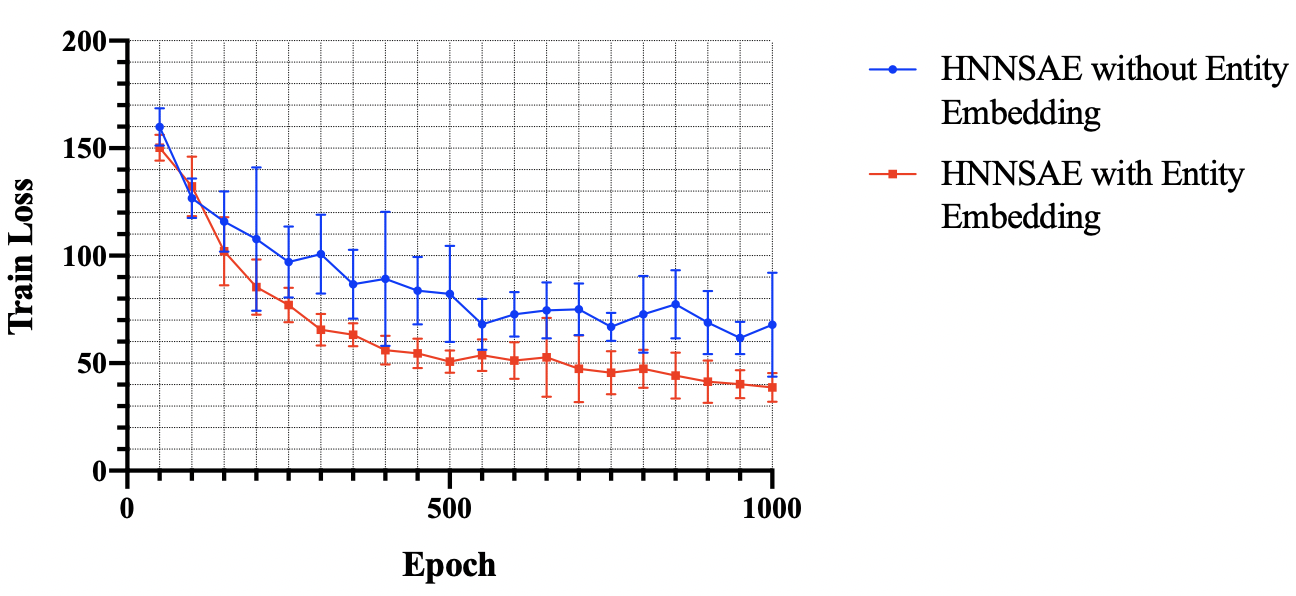}
    \caption{Train loss for model with or without Entity Embedding}
    \label{figure9}
\end{figure}

\begin{figure}[htp]
    \centering
    \includegraphics[width=0.9\linewidth]{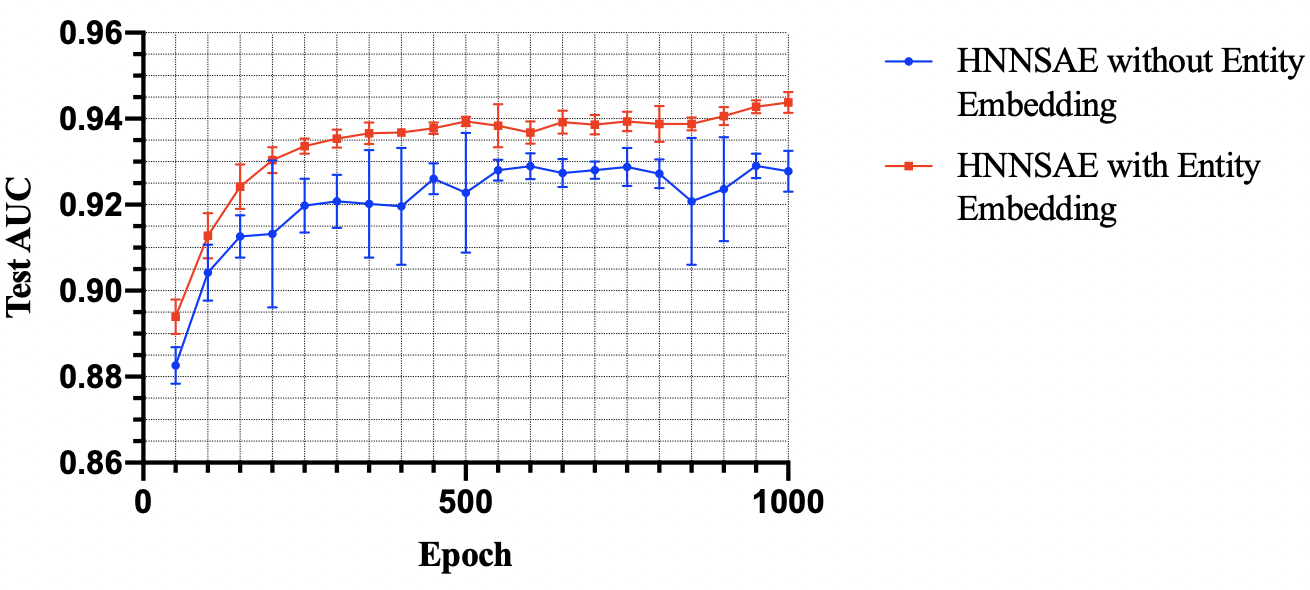}
    \caption{Test AUC for model with or without Entity Embedding}
    \label{figure10}
\end{figure}

\begin{figure}[htp]
    \centering
    \includegraphics[width=0.7\linewidth]{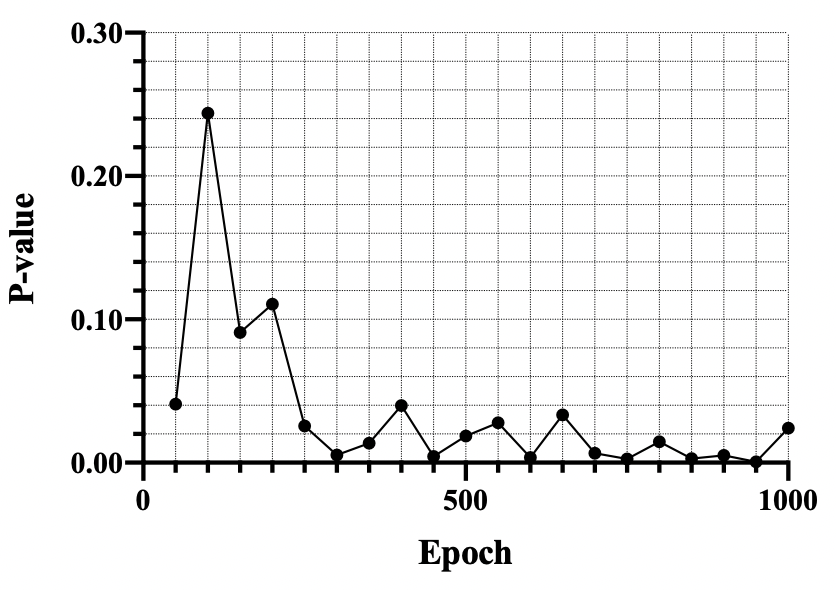}
    \caption{Hypothesis Testing Result(p-value) of Train Loss}
    \label{figure11}
\end{figure}

\begin{figure}[htp]
    \centering
    \includegraphics[width=0.7\linewidth]{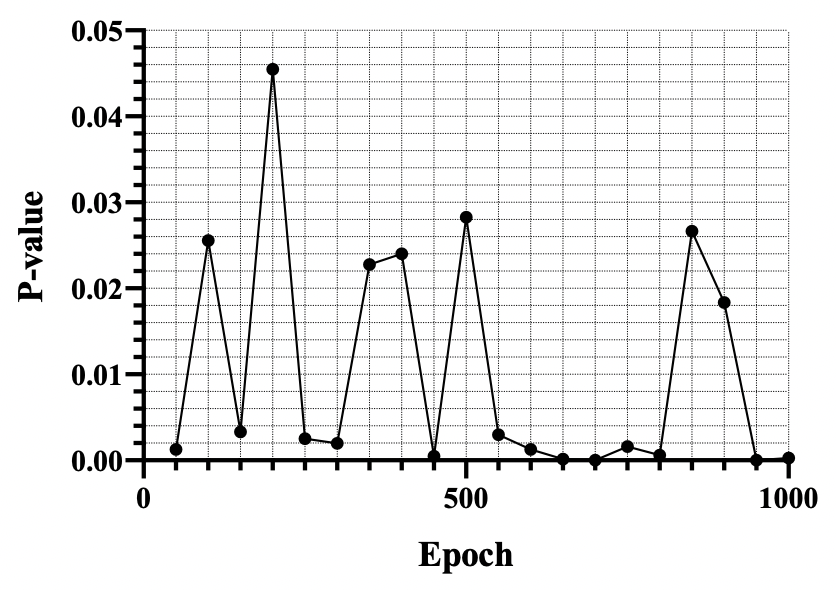}
    \caption{Hypothesis Testing Result(p-value) of Test AUC}
    \label{figure12}
\end{figure}

This section mainly validates the effectiveness of entity embedding in improving the performance of HNNSAE. At the same time, we use hypothesis testing method to verify the significance of entity embedding in improving model performance with the change of epoch. Figure ~\ref{figure9} and ~\ref{figure10} intuitively shows the difference between train loss and test AUC of the two ablation experiment models. Figure ~\ref{figure11} and ~\ref{figure12} respectively shows the p-values of train loss and test AUC hypothesis test of two sets of ablation experiments with every 50 epochs.

From Figure ~\ref{figure9} and Figure ~\ref{figure10}, we can intuitively see that HNNSAE with entity embedding has significantly better predictive performance than HNNSAE without entity embedding. In addition, in terms of the numerical fluctuation range and the overall curve trend of train loss and test AUC, train loss and test AUC of HNNSAE with entity embedding fluctuate less, and in terms of the whole training process, train loss curve and test AUC curve of HNNSAE are relatively smooth. These facts indicate that the introduction of entity embedding improves the stability of model prediction.

Figure ~\ref{figure11} shows that when the number of training epoch is small, the effect of entity embedding on HNNSAE’s train loss reduction is relatively insignificant. However, with the increase of epochs number, p-value shows an obvious downward trend, and the p-value is stable below 0.05 after the 500th epoch, indicating that when the training epoch number is large, entity embedding tends to reduce train loss in HNNSAE more significantly.

It can be observed from Figure ~\ref{figure12} that the p-value of average test AUC hypothesis test are all below 0.05 in the whole training process. The fluctuation of p-value is obvious, but with the increase of the number of epochs, the occurrence frequency of lower p-value increases significantly. Therefore, we can also assume that as the number of epochs increases, the entity embedding has a significant effect on the test AUC of HNNSAE. Entity embedding has a significant on the predictive performance of HNNSAE. 
\subsection{Compare Test AUC with Baselines}
\label{5.5}
To verify the validity of HNNSAE, we conduct experiments from two perspectives. First, HNNSAE is compared with the classical classification baselines. In the individual machine learning model, we choose logistic regression (LR), artificial neural network (ANN) and decision tree (DT) as baselines. The hyperparameter setting of the artificial neural network is consistent with the classifier block in HNNSAE. In the ensemble machine learning model, we choose XGBOOST, Gradient Boosting Decision Tree (GBDT), and Random Forest (RF) as baselines. In addition, in order to verify the effectiveness of the feature extractor based on multi-head self-attention mechanism proposed in this paper, we conducted an ablation experiment, and set the feature extractor as GBDT, Deep Factorization Machine (DeepFM)\citep{guo2017deepfm}, and autoencoder (AE)\citep{zhai2018autoencoder} control group, respectively, on the premise of keeping the hyperparameters of the classifier unchanged. The experimental results are shown in the table below.
\begin{table}[htp]
\centering
\caption{AUC-measure of different methods applied to the commercial bank customer dataset.}
\label{table10}
\begin{tabular}{cc}
\hline
Model             & Test AUC ($mean \pm std$)  \\ \hline
LR                & $0.680 \pm 0.008$          \\
DT                & $0.842 \pm 0.002$          \\
XGBOOST           & $0.905 \pm 0.004$        \\
RF                & $0.897 \pm 0.004$          \\
GBDT              & $0.902 \pm 0.006$          \\
ANN               & $0.779 \pm 0.007$         \\
DeepFM + ANN      & $0.932 \pm 0.008$         \\
GBDT + ANN        & $0.929 \pm 0.006$          \\
AutoEncoder + ANN & $0.798 \pm 0.005$          \\
\textbf{HNNSAE}   & \textbf{$0.944 \pm 0.002$} \\ \hline
\end{tabular}
\end{table}

As can be seen from Table 10, the average test AUC of the five experiments of HNNSAE is 0.9438, which is significantly higher than the baselines of IML, EML and DL selected in this paper. Based on the standard deviation of the test AUC, HNNSAE is more robust than the baselines. The feature filter in HNNSAE effectively extract the significant features and ANN extracts more high-level features. In addition, we find that the feature extraction effect of multi-attention self-attention mechanism is significantly better than DeepFM, GBDT and AutoEncoder. The model proposed in this paper combines the advantages of self-attention and ANN effectively, which may be the reason why HNNSAE can outperform baselines significantly. 

\section{Conclusions and Discussion}
As an important part of customer relationship management, customer churn prediction directly affects the effectiveness and timeliness of customer retention strategy formulation. High-performance customer churn models take proactive actions to reduce the risk of profit and reputation loss for companies. Therefore, the construction of a high accuracy customer churn prediction model has attracted more and more researchers and practitioners’ attention. Artificial feature engineering cannot always extract features effectively, which is one of the reasons that the performance of customer churn prediction model encounters the bottleneck. This paper proposes a hybrid neural network model (HNNSAE), which combines the advantages of self-attention and ANN effectively. A feature filter is constructed based on multi-head self-attention, which adaptively screen the features of high importance. High-level features are extracted by stacking hidden layers after feature filter delivers the output to ANN to achieve better classification effect. The experiment results show that the HNNSAE has a maximum test AUC of 0.947, far outperforming the well-known model commonly adopted in the customer churn prediction community. In addition, through a large number of ablation experiments, we found that in the data set used in this paper, (1) Sample imbalance and redundant attention heads significantly increased the risk of over-fitting of the model. (2) The interaction between the number of attention heads and whether SMOTE is used to handle sample imbalance will significantly affect the predictive performance of the model and the risk of overfitting. (3) The introduction of entity embedding effectively improves model performance, and this improvement tends to be significant as the number of epochs increases. In terms of the limitations of the model, one challenge of HNNSAE is model efficiency. Self-attention needs to compute the dot product between two tokens, so it has high time and space complexity. The impact of this problem will become obvious with the increase of data volume. Therefore, future research will focus on improving the efficiency of the feature filter module of the HNNSAE model. For example, some lightweight attention (e.g. sparse attention variants) can be adopted as a new feature extractor to improve the efficiency of the model\citep{lin2021survey}. Ideally, future improved models should pursue a trade-off between precision and efficiency.

\bibliography{sample}

\end{document}